\documentclass{article}

\usepackage{arxiv}

\usepackage[utf8]{inputenc} % allow utf-8 input
\usepackage[T1]{fontenc}    % use 8-bit T1 fonts
\usepackage{hyperref}       % hyperlinks
\usepackage{url}            % simple URL typesetting
\usepackage{booktabs}       % professional-quality tables
\usepackage{amsfonts}       % blackboard math symbols
\usepackage{nicefrac}       % compact symbols for 1/2, etc.
\usepackage{microtype}      % microtypography
\usepackage{lipsum}
\usepackage{graphicx}

\usepackage{enumitem}
\usepackage{graphicx}
\usepackage{amsmath}
\usepackage{booktabs}
\usepackage{pifont}
\usepackage{float}
\newcommand{\cmark}{\ding{51}}
\newcommand{\xmark}{\ding{55}}
\usepackage{array}
\newcolumntype{C}[1]{>{\centering\arraybackslash}m{#1}}

\title{Human Activity Recognition using Wearable Sensors: Review, Challenges, Evaluation Benchmark}

\author{

Reem Abdel-Salam \thanks{Equal Contribution}\\
  Department of Computer Engineering\\
  Cairo University\\
  \texttt{reem.abdelsalam13@gmail.com} \\

   \And
    
  Rana Mostafa \footnotemark[1]\\
  Department of Computer Engineering\\
  Cairo University\\
  \texttt{ranamostafamohsen96@yahoo.com} \\
  
   \And
  Mayada Hadhood \\
  Department of Computer Engineering\\
  Cairo University\\
  \texttt{mayada.hadhoud@eng.cu.edu.eg} \\
}

\begin{document}
\maketitle

\begin{abstract}
Recognizing human activity plays a significant role in the advancements of human-interaction applications in healthcare, personal fitness, and smart devices. Many papers presented various techniques for human activity representation that resulted in distinguishable progress. In this study, we conduct an extensive literature review on recent, top-performing techniques in human activity recognition based on wearable sensors. Due to the lack of standardized evaluation and to assess and ensure a fair comparison between the state-of-the-art techniques, we applied a standardized evaluation benchmark on the state-of-the-art techniques using six publicly available data-sets: MHealth, USCHAD, UTD-MHAD, WISDM, WHARF, and OPPORTUNITY. Also, we propose an experimental, improved approach that is a hybrid of enhanced handcrafted features and a neural network architecture which outperformed top-performing techniques with the same standardized evaluation benchmark applied concerning MHealth, USCHAD, UTD-MHAD data-sets. 
\end{abstract}

% keywords can be removed

\section{Introduction}
Human Activity Recognition (HAR) is a challenging problem that targets to predict human gestures through computer interaction. It facilitates human lives through a various number of applications. There are two main approaches for human activity recognition:  video images-based recognition and wearable sensors-based recognition. Recognizing human activity from video systems relies on the camera. Not only does this approach require expensive infrastructure installations for cameras, but it also poses some challenges due to background, lighting, and scaling conditions that would lead to difficulty in motion detection. As for the second approach, human activity detection based on wearable sensors such as barometers, accelerometers, gyro-meter, etc.. transforms motion into identified signals. It offers an alternative way to acquire motion without suffering from the same environmental constraints as in the video-based approach as well as offering privacy for its users. However, activity recognition based on this approach has some limitations regarding obtaining sufficient information about all pose movements in the human body that may affect the performance negatively. It is preferable in industrial applications to use more than one input sensor for recording human gestures more accurately and boosting performance.

The focus of this study is directed towards human activity techniques based on wearable sensors. Although there were remarkable improvements in this approach, it is difficult to assess the quality of work in this field due to the lack of standardized evaluation. Our work is reflected in the following contributions:
\begin{enumerate}[itemsep=0pt]
\item Extensive Literature review for recent, top-performing techniques in human activity based on sensor data
\item Due to different evaluation methodologies, it is hard to achieve a fair comparison between recent techniques. Therefore, we applied a standardized evaluation benchmark on the recent works using six publicly available datasets with 3 different temporal windows techniques: Full-Non-Overlapping, Semi-Non-Overlapping, and Leave-One-Trial-Out.
\item Implementation, training, and re-evaluation of the recent literature work using the proposed standardized evaluation benchmark so all techniques follow the same experimental setup to ensure a fair comparison.
\item Proposal of an experimental, hybrid approach that combines enhanced feature extraction with neural networks, and evaluation using the proposed evaluation benchmark criteria, achieving a competitive accuracy.
\end{enumerate}

The rest of the paper is organized as follows: in section 2, data-sets used are demonstrated. In section 3, an extensive literature review of human activity recognition using wearable sensors is discussed further in detail. In section 4, our proposed hybrid approach is introduced. In section 5, the experimental evaluation of human activity recognition for several top works is addressed. Conclusion and recommendations for future work are outlined in section 6.
\section{Data-sets}
\label{DataSets}
Datasets in HAR consist of two main types: Vision-based datasets and Sensor-based datasets. Examples of Vision-based datasets are KTH \cite{KTH} and Wieszmann \cite{Weizmann}. The Sensor-based datasets involve four types: Object sensors, Body-Worn sensors, Hybrid sensors, and Ambient sensors. Vankastern Benchmark \cite{Kasteren2011HumanAR} and Ambient kitchen \cite{Saeed2019MultitaskSL} are examples of datasets based on Object sensors, UCI-HAR and WISDM \cite{Dua:2019} are datasets for Body-worn sensors, Opportunity \cite{Dua:2019} is a dataset based on Hybrid sensors, and the AAL dataset \cite{Anguita2013APD} is for Ambient sensors. We will focus mainly on this study on wearable sensors data-sets. For human activity recognition based on wearable sensor data, there are several open-source data-sets available that offer diversity in categories such as the number of activities to be classified, the number of sensors used, and the sampling rate. In our study, we conduct our experiments on data-sets mentioned below:
\begin{enumerate}[itemsep=0pt]
\item MHealth \cite{Dua:2019}
\item USC-HAD  \cite{Zhang2012USCHADAD}
\item UTD-MHAD \cite{Chen2015UTDMHADAM} 
\item WISDM \cite{Dua:2019}
\item WHARF  \cite{WHARF}
\item OPPORTUNITY \cite{Dua:2019}
\end{enumerate}
\textbf{MHealth} dataset was collected from 10 volunteers performing 12 physical activities such as standing, sitting and relaxing, lying down, walking, climbing stairs, etc. The readings were collected using three sensors: magnetometer, gyroscope, acceleration. The sensors were placed on the chest, right wrist, and left ankle. All the dataset activities are balanced except for the last activity. \textbf{USC-HAD} dataset consists of 12 daily life activities: walk forward, walk left, walk right, walk upstairs, walk downstairs, etc. Those activities were carried out by 14 volunteers. The readings were collected using MotionNode, which integrates a 3-axis accelerometer, a 3-axis gyroscope, and a 3-axis magnetometer. MotionNode sensor was placed on the volunteer's front right hip. The challenge in this dataset that the sensors may not align their readings due to their different sensors' placement. Each sensor responds differently to the human activity performed. \textbf{UTD-MHAD} dataset consists of 27 actions (controlled condition actions): swipe left, swipe right, wave, clap, throw, arms cross, basketball shoot, draw X, draw a circle (clockwise), draw circle (counter-clockwise), draw triangle, etc. These activities were carried out by 8 volunteers. The dataset was collected using a Microsoft Kinect sensor and a wearable inertial sensor in an indoor environment. The inertial sensor was worn on the volunteer's right wrist or the right thigh. \textbf{WISDM} dataset consists of 18 daily living activities classified into Non-hand-oriented activities, Hand-oriented activities, and Hand-oriented activities. The activities include walking, jogging, ascending and descending the stairs, sitting, standing, kicking a ball, etc. Those activities were carried out by 51 volunteers. The dataset was collected using the accelerometer and gyroscope data from both the smartphone placed on the right pants pocket of the volunteer and the smartwatch placed on the volunteer's dominant hand. \textbf{WHARF} dataset consists of 14 daily life activities classified into five categories: toileting, transferring, feeding, ability to use a telephone, and indoor transportation. Those activities - brush own teeth, comb own hair, get up from the bed, lie down on the bed, sit down on a chair, stand up from a chair, drink from a glass, eat with a fork and knife, eat with a spoon, pour water into a glass, use the telephone, climb the stairs, descend the stairs, and walk - were carried out by 17 volunteers. The dataset was collected using an ad hoc sensing device that contains an accelerometer, worn at the right wrist. \textbf{OPPORTUNITY} dataset consists of 21 daily life activities classified into two types locomotion and hand gesture. These activities were carried out by 12 volunteers. The activities include standing, opening the dishwasher, opening drawer1, opening drawer2, opening drawer3, sitting, closing the dishwasher, closing drawer1, closing drawer2, closing drawer3, walking, etc. The dataset was collected using 72 sensors of 10 modalities, integrated with the environment, in objects, and on the body. The sensors include 24 custom Bluetooth wireless accelerometers and gyroscopes, 2 Sun SPOTs and 2 InertiaCube3, the Ubisense localization system, and a custom-made magnetic field sensor. The five X-sense inertial measurement units are placed on a custom-made motion jacket, 12 Bluetooth 3-axis acceleration sensors on the limbs, and commercial InertiaCube3 inertial sensors located on each foot. OPPORTUNITY poses challenges since it is an unbalanced, multi-modal composite dataset.
\subsection{Data-set Preparation}
\label{DataSetsPreparation}
Before feeding data to the model, raw data needs to be transformed into identified samples. In this generation process, data split into equally-sized small windows - or another term temporal windows. Temporal windows are then divided into train and test data sets. Temporal windows may suffer from overlapping in part of the window that may lead to a non-bias evaluation. There are three techniques to generate temporal windows - as mentioned in \cite{Jordao2018HumanAR} - for fair experimental evaluation:
\begin{enumerate}[itemsep=2pt]
\item Full-Non-Overlapping Window represents a generation technique to assure zero overlaps between temporal windows.
\item Semi-Non-Overlapping Window is an alternative approach to Full-Non-Over-lapping-Window for sample generation with a 50\% overlap between every temporal window. This approach will generate a high number of samples, unlike the Full-Non-Over-lapping approach. However, it will result in biased results since the overlapping content may be seen in training and testing.
\item Leave-One-Trial-Out is a novel approach for sample generation as discussed in \cite{Jordao2018HumanAR}. The trial represents a raw activity signal per single subject. It guarantees a non-biased evaluation and sufficient sample number generation. In this generation technique, trials with the same raw signals are not duplicated in training and testing data sets.
\end{enumerate}

Table \ref{tab:SuppTempWindowTechDataSets} provides a checklist of temporal window generation techniques supported in each of the six open-source data-sets discussed above. For the UTD-MAHD dataset, it has been separated into two partitions UTD-1 and UTD-2 based on sensor position. Based on supported temporal window variants of our data-sets, we conduct our experimental results.
\begin{table}
\centering
 \caption{Checklist of Supported Temporal Window Generation Technique for datasets}
  \begin{tabular}{C{2.3cm}C{2.5cm}C{2.5cm}C{2.5cm}} 
    \toprule
    \cmidrule(r){1-4}
    Data-set  & Full-Non-Overlapping Window & Semi-Non-Overlapping Window & Leave-One-Trial-Out\\
    \midrule
    MHealth    &\cmark&\cmark&\cmark \\
    USCHAD     &\cmark&\cmark&\cmark \\
    UTD-1      &\cmark&\cmark&\cmark \\
    UTD-2      &\cmark&\cmark&\cmark \\
    WHARF      &\cmark&\cmark&\cmark \\
    WISDM      &\cmark&\cmark&\cmark \\
    OPPORTUNITY &\xmark&\cmark&\xmark \\
    \bottomrule
  \end{tabular}
  \label{tab:SuppTempWindowTechDataSets}
\end{table}

\section{Literature Review}
\label{LiteratureReview}
HAR is concerned with the ability to understand human behavior. Various approaches to recognize activities have been addressed. The aim is to build a model that predicts the activity sequence based on sensors reading. A considerable amount of literature has been published in human activity recognition based on wearable sensors over the past few years. The next sub-section will provide a comprehensive review related to different approaches for HAR classification.
\subsection{Hand Crafted Methods}
In handcrafted methods, methodology usually starts with extracting important features from the dataset, then applying a classical machine learning technique instead of using deep learning to do both. \textbf{Kwapisz et al.} \cite{Kwapisz2011ActivityRU} worked on \textbf{WISDM} dataset, and extracted features per sensor reading. The authors analyzed three classifiers - Multi-Layer Perception, J48, and logistic regression - to determine the best classifier. MLP out-performed the rest of the classifiers and \textbf{achieved 91.7\%}. \textbf{Catal et al.} \cite{Catal2015OnTU} proposed to use a voting classifier between J48, Logistic Regression, and Multi-Layer Perceptron with the same dataset and set of features as \cite{Kwapisz2011ActivityRU}. Confusion matrix, AUC, F-measure, and accuracy per each category in the data-set were used for evaluation. 
It can be inferred that the features introduced by \cite{Kwapisz2011ActivityRU}: Average-A, Standard Deviation-SD, Average Absolute Difference-AAD, Average Resultant Acceleration-ARA, are the best representative features for sensor-based HAR since they achieve almost the same accuracies compared to deep learning models.

\subsection{CNN based Methods}
\textbf{Ha and Choi} \cite{Ha2016ConvolutionalNN} introduced 2 CNN models: CNN-pf and CNN-pff. CNN-pf represents CNN models with partial weight sharing in the first convolutional layer and full weight sharing in the second convolutional layer. Meanwhile, CNN-pff represents CNN models with partial and full weight sharing in the first convolutional layer, and full weight sharing in the second convolutional layer. The authors used the MHealth dataset using the Semi-Non-Overlapping window, and Leave-One-Subject-Out. The mean accuracy for \textbf {CNN-pf was 91.33\%}, and \textbf{CNN-pff was 91.94\%}. It can be inferred that CNN-pff achieves higher accuracy since they apply partial and full weight sharing. The first convolutional layer tends to capture high-end features that help in boosting the recognition. This approach uses 2D convolution, which has fewer model parameters than applying 1D convolution. \textbf{Panwar et al.} \cite{Panwar2017CNNBA} involved five subjects (humans), especially three arm movements (activities). The authors applied three different pre-processing techniques. Each technique results in a different dataset that is used for training. Two architectures were introduced: the first architecture is for the first two types of pre-processing, and the second architecture is for the third technique. Three validation schemes used: 1) Cross-validation evaluation 2) Leave-One-Subject-Out, 3) Hybrid evaluation - out of 4 repetitions, training with 3 sets of data taken from each subject, and testing with the remaining one set from each subject. The best Accuracy 99.8\% was reported by Synthetic data using Cross-Validation. In the paper, there was limited information about Synthetic data. Since it is a private dataset, there might be a likelihood that the model is overfitting. The approach was simple compared to other HAR CNN-based approaches, which indicate that the datasets used were not challenging enough. \textbf{Kasnesis et al.} \cite{Kasnesis2018PerceptionNetAD} proposed Perception-Net. It consists of a Deep Convolutional Neural Network (CNN) that applies a late 2D convolution to multimodal time-series sensor data for efficient feature extraction. The datasets used to evaluate Perception-Net was \textbf{UCL and PAMAP2}.  Both datasets were normalized. For the UCL dataset, the validation scheme was the Leave-3-Subject-Out approach; meanwhile, for PAMAP2, the Leave-One-Subject-Out validation scheme was used. Model achieved \textbf{  97.25\% and 88.56\% accuracy} for \textbf{UCL and PAMAP2} datasets respectively. \textbf{Bevilacqua et al.} \cite{Bevilacqua2018HumanAR} proposed a CNN network for HAR classification. The author collected an Otago exercise program dataset composed of 16 activities, which is further divided into four categories. The dataset consists of 17 participants. The authors segmented the dataset into small overlapping windows corresponding to roughly 2 seconds of movements. The evaluation was based on the \textbf{F-Score and Confusion matrix for each category}. \textbf{Burns and Whyne} \cite{Burns2020PersonalizedAR} proposed two different models: FCN (Fully Convolutional Network) and PTN (Personalized Feature Classifier). The datasets used for model evaluation were \textbf{WISDM, SPARS, MHealth}. A four-second sliding window was used for MHealth and SPAR, and a ten-second window was utilized for WISDM. The validation scheme used was a 5-fold cross-validation grouping folds by subject. PTN achieved best results on MHealth 99.9\% $\pm$ 0.003, WISDM 91.3\% $\pm$ 0.053, and SPAR 99.0\% $\pm$ 0.017.
\subsection{LSTM-CNN Methods}
\textbf{Lyu et al.} \cite{Lyu2017PrivacyPreservingCD} introduced an LSTM-CNN model for HAR classification, using \textbf{UCI-HAR, Mobile health} dataset with a privacy-preserving scheme for model evaluation. On the UCI-HAR dataset, it was segmented using a fixed sliding window of 2.56 sec and 50\% overlap. For the MHealth dataset, it was segmented in fixed-width sliding windows - 128 readings/windows. The model achieved 95.56\% and 98.44\% on the MHealth dataset and UCI-HAR dataset respectively. The authors use local and dense properties from convolution and learn the temporal structure by storing information in LSTM units by placing a CNN layer above the LSTM layer. This approach achieves better recognition rates for datasets with a high sampling interval since it takes into consideration the activity changes over a time interval. \textbf{Xia et al.} \cite{Xia2020LSTMCNNAF} proposed an LSTM-CNN Model for HAR classification. This model extracts activity features and classifies them with a few model parameters. The authors used \textbf{ UCI-HAR, WISDM, and OPPORTUNITY datsets} with some pre-processing applied. Semi-Non-Overlapping-Window was used to segment the data collected by motion sensors. The validation scheme used was leave-some-subjects-out, depending on participants' number in each. The Evaluation metric was \textbf{F1 score}. The Model achieved \textbf{95.80\% on UCI-HAR}, \textbf{95.75\% on WISDM }, and \textbf{92.63\% on OPPORTUNITY, Gesture recognition}. The approach used the Global Average Pooling layer, which reduces the number of parameters significantly, allows faster convergence of the model, and decreases over-fitting.
\subsection{CNN-LSTM Methods}
\textbf{Sun et al.} \cite{Sun2018SequentialHA} proposed a CNN-LSTM-ELM network and used \textbf{OPPORTUNITY data-set}. Pre-processing techniques are applied to overcome challenges in the OPPORTUNITY data-set. OPPORTUNITY data-set activities can be divided into Gesture and Locomotion. The authors used the Gesture category in the OPPORTUNITY data-set. \textbf{F1 and accuracy} were used as a performance measure for the model. Experiments were conducted on the CNN-LSTM-ELM network and CNN-LSTM Fully connected network. It was found out that the ELM classifier is generalizing better and faster than Fully connected. The CNN-LSTM-ELM model achieved \textbf{91.8\% accuracy for gesture recognition}, while CNN-LSTM-Fully connected model achieved \textbf{89.7\% accuracy for gesture recognition}. \textbf{Wang et al.}  \cite{Huaijunetal} proposed a 1D CNN-LSTM network to learn local features and model the time dependence between features. The model consists of 3 alternating 1D convolution layer and max pool followed by LSTM layer, fully connected layer, and batch normalization. The authors used  \textbf{international standard Data Set, Smartphone-Based Recognition of Human Activities and Postural Transitions Data Set (HAPT)}\cite{Abidine2016TheJU,Weiss2016ActitrackerAS}. The HAPT data set contains twelve types of actions. These actions can be classified into 3 types: static, walking, and transitions between any two static movements. The authors experimented using different model CNN, LSTM, CNN-GRU, CNN Bi-LSTM, CNN-LSTM. The best performing model was the CNN-LSTM, which achieved \textbf{95.87\%} accuracy on the dataset.\\ \\
Based on the recent works' findings, we conclude that the CNN-LSTM and LSTM-CNN techniques achieve overall higher accuracy. We believe that is a result of using LSTM layers as they take into account the history of the signals, which may lead to better recognition. The drawbacks of using such techniques that they take more computational power and time for training. The advantage of using LSTM-CNN techniques for human activity recognition is that the LSTM captures time dependencies first. Then, the features are extracted based on the time dependencies using CNN. However, using CNN-LSTM techniques, the features are captured first, then considering the sequence of the features in time. This methodology may not align well with time dependencies; thus, it will make convergence harder with more computational power needed. It can be inferred that the recent works use a different experimental setup, each proposing its own evaluation benchmark. Some wearable data-sets are private, so it makes it hard to reproduce the results and conduct a fair comparison based on the work findings. Due to the absence of a unified evaluation criteria on all recent works, we applied a standardized evaluation benchmark to perform a fair, non-biased performance comparison. We compare between recent works concerning the standardized benchmark in \ref{Results}. 
\section{Proposed Hybrid Approach}
This section describes our proposed hybrid approach in detail. Our methodology is divided into two stages: enhanced feature extraction followed by Neural Network architecture. 
\subsection{Feature Extraction}
\label{featureExtraction}
Before feeding our data to our NN architecture, features were extracted from sampled data. We used 12 features - 4 of them were proposed by \cite{Kwapisz2011ActivityRU}. In summary, the features are Average-A, Standard Deviation-SD, Average Absolute Difference-AAD, Average Resultant Acceleration-ARA, Maximum, Minimum, Median, Skew, Kurtosis, Interquartile range, Area under the curve, and Square area under the curve.
After the extraction of features, scale normalization, and Principal Component Analysis (PCA) \cite{PCAAnalysis} is further applied to the extracted features to remove redundant features -  due to window overlapping - that helped in boosting accuracy.
\subsection{NN Architecture}
Our proposed architecture consists of three dense layers followed by softmax of activity categories' number in the dataset. Our architecture consists of a fully connected layer of size 128, followed by another fully connected layer of size 64 and then followed by 32 fully-connected layers. We used Adam as an optimizer with batch size 16 and Leaky-Relu as an activation. 

The 12 Features are computed per each sample window in the dataset. Then, our lightweight neural network trains using the extracted features as input. The neural network learns the hidden features and optimizes its weights to reach a higher recognition accuracy than other classical approaches that use classical machine learning techniques.

Our proposed hybrid approach is stable and lightweight, compared with other techniques. It has demonstrated high competitive accuracy for HAR datasets as well as it can be easily deployed on resource-constrained hardware. Last but not least, it is found to be compatible with HAR datasets with no modifications needed.
\section{Experimental Results}
This section outlines our experiments and performance comparison between different approaches under the same evaluation criteria. We conducted several experiments with respect to our standardized evaluation benchmark. The evaluation metric is discussed in detail below.

\subsection{Evaluation Metric}
Recent approaches discussed in section \ref{LiteratureReview} are implemented\footnote{Recent works are implemented using the same architecture and hyper-parameters as mentioned in their papers and re-evaluated using proposed standardized benchmark}, trained and re-evaluated alongside our hybrid approach to follow the same experimental setup using a standardized benchmark: 6 publicly available data-sets and 3 temporal window techniques described in \ref{DataSetsPreparation}. We conducted our experiments using Google colab with 1xTesla T4 GPU, 2496 CUDA cores, and 12GB GDDR5 VRAM. In our study, mean accuracy is taken into account as an evaluation criterion for results.

\subsection{Experimental Set-Up}
\label{ExperimentResults}
We conducted three kinds of experiments based on the validation technique used: \\\\
\textbf{K-Folds Validation Experiment}
\label{KFolds}
For the first experiment, top-performing approaches alongside our proposed approach are being evaluated via the K-Folds validation technique. We conduct our experiment using:  MHealth, USCHAD, UTD-1, UTD-2, WHARF, and WISDM data-sets, concerning 3 generation techniques discussed earlier in section \ref{DataSets}.\\
\textbf{Leave-One-Subject-Out Experiment} 
In this experiment, the same set-up is used as in the K-Folds experiment. The difference in this experiment is that the evaluation is conducted via the Leave-One-Subject-Out validation technique for the Semi-Non-Overlapping-Window sample generation technique only.\\ 
\textbf{Hold-Out Validation Experiment}
In this experiment, we investigated the effect of different hyper-parameters on our proposed approach via Hold-Out validation. Two variants of our proposed model were introduced: Proposed Approach V1 and Proposed Approach V2. Proposed Approach V1 training was set for 250 epochs; meanwhile, Proposed Approach V2 was set for 200 epochs. This experiment is divided into two separate trials:
\begin{enumerate}
    \item The first is to compare the two variants of our proposed approach alongside other top-performing methods and evaluate accuracies using the OPPORTUNITY data-set and Semi-Non-Overlapping-Window technique
    \item The second is to report the accuracy of two variants of our proposed approach: Proposed Approach V1 and Proposed Approach V2 using: MHealth, USCHAD, UTD-1, UTD-2, WHARF, and WISDM data-sets, with respect to 3 temporal window sample generation techniques.
\end{enumerate}

\subsection{Results}
\label{Results}
In this section, we conduct experiments mentioned in \ref{ExperimentResults} to evaluate recent state-of-the-art approaches, alongside with our proposed method. Top accuracies and second-top accuracies are highlighted with bold and underlined respectively. 

Firstly, we discuss the results of the K-Folds Validation technique experiment. Mean accuracies using Semi-Non-Overlapping-Window and K-Folds Validation are reported in Table ~\ref{tab:SNOW}. It can be inferred that our approach ranked top accuracy in MHealth, USCHAD, UTD-1, and UTD-2 datasets. For the WHARF dataset, Lyu et al. \cite{Lyu2017PrivacyPreservingCD} achieved top accuracy of 88.99\%. Xia et al. \cite{Xia2020LSTMCNNAF} ranked top accuracy with a 1.3\% accuracy difference relative to our approach in the WISDM dataset. 

For Leave-One-Trial-Window and K-Folds Validation experiment demonstrated in Table ~\ref{tab:LOTO}, our technique ranked top accuracy in MHealth, USHCAD, UTD-1, UTD-2 datasets. Xia et al. \cite{Xia2020LSTMCNNAF} achieved top accuracy of 91.02\% with a 5\% increase in accuracy compared to our method for the WISDM dataset. Lyu et al. \cite{Lyu2017PrivacyPreservingCD} obtained top accuracy in the WHARF dataset.

Referring to the Full-Non-Overlapping-Window reported in Table ~\ref{tab:FNOW}, our approach ranked top accuracy for MHealth, USCHAD, UTD-1, and UTD-2 datasets. For the WHARF and WISDM dataset, our approach ranks third-best accuracy.

Based on the results of experiments reported above for Tables \ref{tab:SNOW}, \ref{tab:LOTO} and \ref{tab:FNOW}, it can be concluded that our proposed approach outperforms state-of-the-art techniques - both conventional and deep learning techniques - for MHealth, USCHAD, UTD-1, and UTD-2 datasets for 3 window generation techniques respectively. For WISDM and WHARF, our approach, in most of the trials, has ranked as one of the top three best accuracies. We believe that the reason behind the drop in our proposed approach's accuracy for WISDM and WHARF data-sets is the low sampling rate and that only one sensor was used for signal readings.

\begin{table}
 \caption{Mean Accuracy using Semi-Non-Overlapping-Window and K-Folds Validation. (-) denotes that the approach is incompatible with the dataset and window technique used}
  \centering
  \begin{tabular}{ccccccc}
    \toprule
    \cmidrule(r){1-6}
    Approach&MHealth &USCHAD &UTD-1 &UTD-2 &WHARF &WISDM \\
    \midrule
    Bevilacqua et al. \cite{Bevilacqua2018HumanAR} &93.11&-&-&-&-&-\\
    Catal et al.\cite{Catal2015OnTU}        &99.84&\underline{91.18}&49.06&81.07&66.42&90.60 \\
    Burns and Whyne \cite{Burns2020PersonalizedAR} &95.54&-&33.39&69.05&62.40&98.82 \\
    Ha and Choi \cite{Ha2016ConvolutionalNN} &84.77&-&22.67&61.67&68.95&81.81 \\
    Xia et al. \cite{Xia2020LSTMCNNAF}  &\underline{99.96}&-&56.41&84.62&\underline{87.45}&\textbf{99.65} \\
    Sun et al. \cite{Sun2018SequentialHA} (Fully Connected) &-&-&-&-&-&-  \\
    Sun et al. \cite{Sun2018SequentialHA} (ELM) &83.34&-&23.63&56.06&57.06&- \\ 
    Kasnesis et al. \cite{Kasnesis2018PerceptionNetAD} &12.45&39.48&6.85&15.82&-&-\\
    Lyu et al. \cite{Lyu2017PrivacyPreservingCD} &99.77&-&\underline{61.53}&\underline{86.13}&\textbf{88.99}&\underline{99.47} \\
    Panwar et al. \cite{Panwar2017CNNBA} &09.00&13.84&05.22&51.59&-&-  \\
    Proposed Approach    &\textbf{100}&\textbf{93.48}&\textbf{71.62}&\textbf{87.98}&80.39&98.35 \\
    \bottomrule
  \end{tabular}
  \label{tab:SNOW}
\end{table}
\begin{table}
 \caption{Mean Accuracy using Leave-One-Trial-Window and K-Folds Validation}
  \centering
  \begin{tabular}{ccccccc}
    \toprule
    \cmidrule(r){1-6}
    Approach      &MHealth &USCHAD &UTD-1 &UTD-2 &WHARF &WISDM \\
    \midrule
    Bevilacqua et al. \cite{Bevilacqua2018HumanAR} &89.62&-&-&-&-&-\\
    Catal et al. \cite{Catal2015OnTU}       &\underline{91.76} &\underline{87.36}&47.96&80.35&63.87&80.11 \\
    Burns and Whyne \cite{Burns2020PersonalizedAR} &89.77&-&33.64&69.02&61.59&- \\
    Ha and Choi \cite{Ha2016ConvolutionalNN} &76.66&-&21.46&63.87&64.68&76.42 \\
    Xia et al. \cite{Xia2020LSTMCNNAF}  &87.89 &90.94&53.97&82.44&\underline{83.47}&\textbf{91.02} \\
    Sun et al. \cite{Sun2018SequentialHA} &77.69&-&24.30&-&-&-  \\
    Sun et al. \cite{Sun2018SequentialHA} (ELM) &80.27&-&27.46&48.44&61.92&- \\ 
    Kasnesis et al. \cite{Kasnesis2018PerceptionNetAD}   &14.33 &40.61&6.39&18.13&-&- \\
    Lyu et al. \cite{Lyu2017PrivacyPreservingCD} &89.17&-&\underline{57.10}&\underline{83.64}&\textbf{85.07}&\underline{89.19}\\
    Panwar et al. \cite{Panwar2017CNNBA} &09.02&13.85&05.24&50.00&-&-  \\
    Proposed Approach   &\textbf{94.76}&\textbf{90.94}&\textbf{71.00}&\textbf{87.18}&\t77.20&86.18 \\
    \bottomrule
  \end{tabular}
  \label{tab:LOTO}
\end{table}
\begin{table}
 \caption{Mean Accuracy using Full-Non-Overlapping-Window and K-Folds Validation}
  \centering
  \begin{tabular}{ccccccc}
    \toprule
    \cmidrule(r){1-6}
    Approach&MHealth &USCHAD &UTD-1 &UTD-2 &WHARF &WISDM \\
    \midrule
    Bevilacqua et al. \cite{Bevilacqua2018HumanAR} &95.96&-&-&-&-&-\\ Catal et al. \cite{Catal2015OnTU}        &99.55&88.79&47.01&\underline{80.32}&60.76&88.84 \\
    Burns and Whyne \cite{Burns2020PersonalizedAR} &93.65&-&26.46&65.55&57.70&96.46 \\
    Ha and Choi \cite{Ha2016ConvolutionalNN} &79.85&-&18.92&57.94&61.70&77.07 \\  Xia et al.\cite{Xia2020LSTMCNNAF}  &\underline{99.70}&\underline{91.55}&45.99&79.76&\underline{78.49}&\textbf{99.06} \\
    Sun et al. \cite{Sun2018SequentialHA} (Fully Connected) &78.77&-&22.70&39.87&44.43&-  \\
    Sun et al. \cite{Sun2018SequentialHA} (ELM) &62.20&-&19.79&37.46&49.18&- \\ 
    Kasnesis et al. \cite{Kasnesis2018PerceptionNetAD}  &9.52&31.74&6.46&16.70&-&- \\
    Lyu et al. \cite{Lyu2017PrivacyPreservingCD} &99.33&89.92&\underline{52.27}&80.21&\textbf{83.29}&\underline{98.46} \\
    Panwar et al. \cite{Panwar2017CNNBA} &08.99&13.50&05.27&48.04&-&-  \\
    Proposed Approach   &\textbf{99.70}&\textbf{91.68}&\textbf{70.48}&\textbf{87.84}&76.02&97.50\\
    \bottomrule
  \end{tabular}
  \label{tab:FNOW}
\end{table}
\begin{table}
 \caption{Mean Accuracy using Semi Non-Overlapping-Window and Leave-One-Subject-Out Validation}
  \centering
  
  \begin{tabular}{ccccccc}
    \toprule
    \cmidrule(r){1-6}
    Approach      &MHealth &USCHAD &UTD-1 &UTD-2 &WHARF &WISDM \\
    \midrule
    Bevilacqua et al. \cite{Bevilacqua2018HumanAR} &85.00 &-&-&-&-&-\\
    Catal et al. \cite{Catal2015OnTU}       &\underline{95.87} &\underline{74.62}&31.98&73.67&49.69&\underline{73.86} \\
    Burns and Whyne \cite{Burns2020PersonalizedAR}  &91.78&-&30.33&64.96&49.87&- \\
    Ha and Choi \cite{Ha2016ConvolutionalNN} &75.69&-&19.49&59.58&59.13&59.28 \\
    Xia et al. \cite{Xia2020LSTMCNNAF} &93.81&-&32.60&71.02&\underline{65.13}&- \\
    Sun et al. \cite{Sun2018SequentialHA} (Fully Connected) &78.48&59.13&18.97&51.17&49.39&-  \\
    Sun et al. \cite{Sun2018SequentialHA} (ELM) &81.57&-&19.37&50.90&50.52&- \\ 
    Kasnesis et al. \cite{Kasnesis2018PerceptionNetAD}   &13.23&44.45&6.68&18.40&-&- \\
    Lyu et al. \cite{Lyu2017PrivacyPreservingCD} &92.05&-&\underline{36.73}&\underline{74.77}&\textbf{70.95}&- \\
    Panwar et al. \cite{Panwar2017CNNBA} &09.01&14.72&05.23&40.63&-&-  \\
    Proposed Approach &\textbf{96.35}&\textbf{74.71}&\textbf{50.82}&\textbf{81.37}&59.29&\textbf{77.91} \\
    \bottomrule
  \end{tabular}
  \label{tab:LOSO}
\end{table}
\begin{table}
 \caption{Mean Accuracy using Opportunity with Semi-Non-Overlapping-Window and Hold-Out Validation}
  \centering
  \begin{tabular}{cc}
    \toprule
    \cmidrule(r){1-2}
    Approach     &Opportunity \\
    \midrule
    Bevilacqua et al. \cite{Bevilacqua2018HumanAR} &\textbf{86.68}\\
    Catal et al. \cite{Catal2015OnTU}  &85.45 \\
    Burns and Whyne \cite{Burns2020PersonalizedAR} &-\\
    Ha and Choi \cite{Ha2016ConvolutionalNN} &- \\
    Xia et al.\cite{Xia2020LSTMCNNAF}  &- \\
    Sun et al.\cite{Sun2018SequentialHA} (Fully connected) &- \\
    Sun et al. \cite{Sun2018SequentialHA} (ELM) &83.21 \\ 
    Kasnesis et al. \cite{Kasnesis2018PerceptionNetAD}  &83.48\\
    Lyu et al. \cite{Lyu2017PrivacyPreservingCD} &- \\
    Panwar et al. \cite{Panwar2017CNNBA} &84.22 \\
    Proposed Approach V1 &\underline{86.24} \\
    Proposed Approach V2   &85.78 \\
    \bottomrule
  \end{tabular}
  \label{tab:Opportunity}
\end{table}
\begin{table}
\caption{Mean Accuracy using Semi-Non-Overlapping-Window and Hold-Out Validation}
\centering
 \begin{tabular}{C{1.5cm}C{2.5cm}C{2.5cm}}
    \toprule
    \cmidrule(r){1-3}
    Dataset & Proposed Approach V1 & Proposed Approach V2 \\
    \midrule
    MHealth & 100 & 100 \\
    USCHAD & 92.72 & 91.83 \\
    UTD-1 & 65.70 & 66.51 \\
    UTD-2 & 85.64 & 82.71 \\
    WHARF & 74.24 & 74.47 \\
    WISDM & 95.57 & 97.18 \\
    \bottomrule
    \end{tabular}
\label{tab:SNOWHV}
\end{table}

Table ~\ref{tab:LOSO} reports the accuracies of various techniques alongside our approach using Semi Non-Overlapping-Window and Leave-One-Subject-Out. Based on results in Table~\ref{tab:LOSO}, it is found out that our proposed approach out-performs state-of-the-art-methods, achieving top-ranked accuracy using MHealth, USCAHD, UTD-1, UTD-2, and WISDM dataset. Referring to the WHARF dataset, Lyu et al. \cite{Lyu2017PrivacyPreservingCD} ranked top accuracy relative to our model with an 11\% accuracy difference.

Table \ref{tab:Opportunity} demonstrates mean accuracy results using the Hold-Out Validation technique on Opportunity Dataset ( Semi-Non-Overlapping). We investigate the effect of hyper-tuning parameters. Compared with other approaches, Proposed Approach V1 that is trained with 250 epochs ranked second best with 86.24\% accuracy with 0.4\% accuracy difference relative to \cite{Bevilacqua2018HumanAR}.
\begin{table}
 \caption{Mean Accuracy using Leave-One-Trial-Out Window and Hold-Out Validation}
  \centering
  \begin{tabular}{C{1.5cm}C{2.5cm}C{2.5cm}}
    \toprule
    \cmidrule(r){1-3}
    Dataset  & Proposed Approach V1 & Proposed Approach V2 \\
    \midrule
    MHealth & 100 & 99.76 \\
    USCHAD & 92.13 & 91.58 \\
    UTD-1 & 68.51 & 65.22 \\
    UTD-2 & 83.51 & 84.84 \\
    WHARF & 75.04 & 74.73 \\
    WISDM & 97.50 & 96.05 \\
    \bottomrule
  \end{tabular}
  \label{tab:LOTOHV}
\end{table}
\begin{table}
 \caption{Mean Accuracy using Full-Non-Overlapping Window and Hold-Out Validation}
  \centering
  \begin{tabular}{C{1.5cm}C{2.5cm}C{2.5cm}}
    \toprule
    \cmidrule(r){1-3}
    Dataset & Proposed Approach V1 & Proposed Approach V2 \\
    \midrule
    MHealth & 100 & 100 \\
    USCHAD & 89.72 & 90.38 \\
    UTD-1 & 62.13 & 61.83 \\
    UTD-2 & 82.84 & 82.35 \\
    WHARF & 69.82 & 67.84 \\
    WISDM & 95.45 & 94.73 \\
    \bottomrule
  \end{tabular}
  \label{tab:FNOWHV}
\end{table}

Tables \ref{tab:SNOWHV}, \ref{tab:LOTOHV}, \ref{tab:FNOWHV} investigates the effect of training our model with different epochs: Proposed Approach V1 and V2 via Hold-out validation. The results are reported via MHealth, USCHAD, UTD-1, UTD-2, WHARF, and WISDM datasets for the three-generation techniques respectively. It can be concluded that there is an overall increase in the performance of the model compared with K-Folds validation results demonstrated in tables \ref{tab:SNOW}, \ref{tab:LOTO}, and \ref{tab:FNOW}. We believe it is a consequence of using the hold-out validation technique, whereas the model could optimize its parameters over the entire training dataset.

In our experiments, we investigated the performance of our proposed approach not only in terms of recognition accuracy but also in terms of time complexity. We analyzed the time taken for our proposed approach - with respect to the standardized benchmark above - to extract the handcrafted features mentioned in \ref{featureExtraction} per sample window. It was observed that the time taken in seconds to extract the features per sample varies ranging from 0.008(s) - 0.03(s). We believe that the MHealth takes longer to extract features due to its long temporal window size.
\section{Conclusion}
In this study, an extensive literature review on recent, top-performing approaches in human activity recognition based on wearable sensors is addressed. Due to the lack of non-standardized evaluation, recent approaches are implemented and re-evaluated using our standardized benchmark with three data sample generation techniques as discussed in \ref{DataSetsPreparation} to follow the same experimental setup for a fair evaluation. Our experiments were conducted via six open-source datasets. A hybrid experimental approach is proposed for human activity recognition. Features are first extracted using our feature engineering and then followed by 3-layered neural network architecture. Our experimental results demonstrate that our proposed hybrid approach has a strong generalization ability with high recognition accuracy, out-performing all state-of-the-art techniques for MHealth, USCHAD, UTD-1, and UTD-2 datasets. 

Future work should investigate the impact of low-sampling-rate and high activity number datasets such as WHARF and WISDM. More features should be added to our feature extraction approach, and further hyper-tune our neural network approach for higher recognition ability for human activity.

\section{Acknowledgments}
We would like to thank Jordao et al. \cite{Jordao2018HumanAR} for sharing datasets: MHealth, USC-HAD, UTD-MHAD, WHARF, and WISDM  that have been segmented by the temporal window generation techniques publicly to the community.

\bibliographystyle{unsrt}  

\bibliography{har}
\end{document}